\documentclass[11pt]{article}
\usepackage{eacl2017}
\usepackage{times}
\usepackage{url}
\usepackage{latexsym}
\usepackage{amsmath}
\usepackage[algoruled,linesnumbered]{algorithm2e}
\usepackage{lipsum}

\newcommand\blfootnote[1]{%
  \begingroup
  \renewcommand\thefootnote{}\footnote{#1}%
  \addtocounter{footnote}{-1}%
  \endgroup
}

\emnlpfinalcopy 
\def\emnlppaperid{***} 

\title{SCDV : Sparse Composite Document Vectors using soft clustering over distributional representations}

\author{\bf{Dheeraj Mekala*} \hspace{15mm} {Vivek Gupta*} \hspace{16mm} {Bhargavi Paranjape}  \hspace{10mm} {Harish Karnick}\\ IIT Kanpur \hspace{20mm} Microsoft Research \hspace{15mm} Microsoft Research \hspace{16mm} IIT Kanpur \\ {\tt dheerajm@iitk.ac.in} \hspace{2mm} {\tt \{t-vigu,t-bhpara\}@microsoft.com}  \hspace{2mm} {\tt hk@iitk.ac.in}}

\date{}

\begin{document}

\maketitle
\begin{abstract}
\blfootnote{*Represents equal contribution}
We present a feature vector formation technique for documents - Sparse Composite Document Vector (SCDV) - which overcomes several shortcomings of the current distributional paragraph vector representations that are widely used for text representation. In SCDV, word embeddings are clustered to capture multiple semantic contexts in which words occur. They are then chained together to form document topic-vectors that can express complex, multi-topic documents. Through extensive experiments on multi-class and multi-label classification tasks, we outperform the previous state-of-the-art method, NTSG
 \cite{liu2015learning}. We also show that SCDV embeddings perform well on heterogeneous tasks like Topic Coherence, context-sensitive Learning and Information Retrieval. Moreover, we achieve significant reduction in training and prediction times compared to other representation methods. SCDV achieves best of both worlds - better performance with lower time and space complexity.
\end{abstract}

\section{Introduction}
Distributed word embeddings represent words as dense, low-dimensional and real-valued vectors that can capture their semantic and syntactic properties. These embeddings are used abundantly by machine learning algorithms in tasks such as text classification and clustering. Traditional bag-of-word models that represent words as indices into a vocabulary don't account for word ordering and long-distance semantic relations. Representations based on neural network language models \cite{mikolov2013linguistic} can overcome these flaws and further reduce the dimensionality of the vectors. However, there is a need to extend word embeddings to entire paragraphs and documents for tasks such as document and short-text classification.

Representing entire documents in a dense, low-dimensional space is a challenge. A simple weighted average of the word embeddings in a large chunk of text ignores word ordering, while a parse tree based combination of embeddings  \cite{socher2013recursive} can only extend to sentences.  \cite{le2014distributed} trains word and paragraph vectors to predict context but shares word-embeddings across paragraphs.  However, words can have different semantic meanings in different contexts. Hence, vectors of two documents that contain the same word in two distinct senses need to account for this distinction for an accurate semantic representation of the documents.  \cite{wang2015},  \cite{liu2015learning} map word embeddings to a latent topic space to capture different senses in which words occur. However, they represent complex documents in the same space as words, reducing their expressive power. These methods are also computationally intensive.

In this work, we propose the Sparse Composite Document Vector(SCDV) representation learning technique to address these challenges and create efficient, accurate and robust semantic representations of large texts for document classification tasks. SCDV combines syntax and semantics learnt by word embedding models together with a latent topic model that can handle different senses of words, thus enhancing the expressive power of document vectors. The topic space is learnt efficiently using a soft clustering technique over embeddings and the final document vectors are made sparse for reduced time and space complexity in tasks that consume these vectors.

The remaining part of the paper is organized as follows. Section 2 discusses related work in document representations. Section 3 introduces and explains SCDV in detail. This is followed by extensive and rigorous experiments together with analysis in section 4 and 5 respectively.

\section{Related Work}

 \cite{le2014distributed} proposed two models for distributional representation of a document, namely, \textit{Distributed Memory Model Paragraph Vectors (PV-DM)}  and \textit{Distributed BoWs paragraph vectors (PV-DBoW)}. In \textit{PV-DM}, the model is learned to predict the next context word using word and paragraph vectors. In \textit{PV-DBoW}, the paragraph vector is directly learned to predict randomly sampled context words. In both models, word vectors are shared across paragraphs. While word vectors capture semantics across different paragraphs of the text, documents vectors are learned over context words generated from the same paragraph and potentially capture only local semantics  \cite{pranjal2015weighted}. Moreover, a paragraph vector is embedded in the same space as word vectors though it can contain multiple topics and words with multiple senses. As a result, doc2vec  \cite{le2014distributed} doesn't perform well on Information Retrieval as described in  \cite{paraanalysis} and  \cite{roy2016representing}. Consequently, we expect a paragraph vector to be embedded in a higher dimensional space. 
 
A paragraph vector also assumes all words contribute equally, both quantitatively (weight) and qualitatively (meaning). They ignore the importance and distinctiveness of a word across all documents \cite{pranjal2015weighted}. Mukerjee et al.  \cite{pranjal2015weighted} proposed idf-weighted averaging of word vectors to form document vectors. This method tries to address the above problem. However, it assumes that all words within a document belong to the same semantic topic. Intuitively, a paragraph often has words originating from several semantically different topics. In fact, Latent Dirichlet Allocation  \cite{Blei:2003} models a document as a distribution of multiple topics. 

These shortcomings are addressed in three novel composite document representations 
called {\em Topical word embedding (TWE-1,TWE-2 and TWE-3)} by  \cite{liu2015learning}. \textit{TWE-1} learns word and topic 
embeddings by considering each topic as a pseudo word and builds the topical 
word embedding for each word-topic assignment. Here, the interaction between a word and the topic to which it is 
assigned is not considered. \textit{TWE-2} learns a topical word 
embedding for each word-topic assignment directly, by considering each word-
topic pair as a pseudo word. Here, the interaction between a word and 
its assigned topic is considered but the vocabulary of pseudo-words blows up. For each word and each topic, \textit{TWE-3} builds 
distinct embeddings for the topic and word and concatenates them for each word-topic assignment. Here, 
the word embeddings are influenced by the corresponding topic embeddings, 
making words in the same topic less discriminative.


 \cite{liu2015learning} proposed an architecture called {\em Neural tensor 
skip-gram model (NTSG-1, NTSG-2, NTSG-3, NTSG-4)}, that learns 
multi-prototype word embeddings and uses a tensor layer to model the 
interaction of words and topics to capture different senses. $NTSG$  outperforms other embedding methods 
like $TWE-1$ on the 20 newsgroup data-set by modeling context-sensitive embeddings in addition to topical-word embeddings.  $LTSG$ \cite{ltsg} builds on $NTSG$ by jointly learning the latent topic space and context-sensitive word embeddings. All three, $TWE$, $NTSG$ and $LTSG$ use $LDA$ and suffer from computational issues like large training time, prediction time and storage space. They also embed document vectors in the same space as terms. Other works that harness topic modeling like $WTM$ \cite{wtm}, $w2v-LDA$ \cite{wtvlda}, $TV+MeanWV$ \cite{tvMeanWV}, $LTSG$ \cite{ltsg}, $Gaussian-LDA$ \cite{gaussianlda}, $Topic2Vec$  \cite{topic2vec},  \cite{lda2vec} and $MvTM$ \cite{mvtm} also suffer from similar issues.

 \cite{vivek} proposed a method to form a composite document vector using 
word embeddings and tf-idf values, called the {\em Bag of Words Vector 
(BoWV)}. In $BoWV$, each document is represented by a vector of dimension 
$D = K*d + K$, where $K$ is the number of clusters and $d$ is the 
dimension of the word embeddings. The core idea behind $BoWV$ is that semantically different 
words belong to different topics and their word vectors should not be 
averaged. Further, $BoWV$ computes inverse cluster frequency of each 
cluster (icf) by averaging the idf values of its member terms to capture the 
importance of words in the corpus. However, $BoWV$ does hard clustering using K-means algorithm, assigning each word to only one cluster or semantic topic but a word can belong to multiple topics. For example, the word \textit{apple} belongs to topic \textit{food} as a fruit, and belongs to topic \textit{Information Technology} as an IT company. Moreover, $BoWV$ is a non-sparse, high dimensional continuous vector and suffers from computational problems like large training time, prediction time and storage requirements.

\section{Sparse Composite Document Vectors} 
\label{SCDV}
In this section, we present the proposed Sparse Composite Document Vector (SCDV) representation as a novel document vector learning algorithm. The feature formation algorithm can be divided into three steps.

\subsection{Word Vector Clustering}
We begin by learning \textit{d} dimensional word vector representations for every word in the vocabulary 
$V$ using the skip-gram algorithm with negative sampling (SGNS) \cite{mikolov2013distributed}. 
We then cluster these word embeddings using the Gaussian Mixture Models(GMM)  \cite{gmm} soft clustering technique. The number of clusters, \textit{K}, to be formed is a parameter of the SCDV model.
By inducing soft clusters, we ensure that each word belongs to every cluster with some probability $P(c_{k}|w
_{i})$. 

\begin{equation*}
p(c_k=1) = \pi_k
\end{equation*}
\begin{equation*}
p(c_k = 1|w) = \frac{\pi_k\mathcal{N}(w|\mu_k, \Sigma_k)}{\Sigma^K_{j=1}\pi_j\mathcal{N}(w|\mu_j,\Sigma_j)}
\end{equation*}

\begin{algorithm}[h!]
    \SetAlgoNoLine
    \KwData{Documents $D_{n}$, n = 1 $\ldots$ N}
    \KwResult{Document vectors $\vec{SCDV_{D_{n}}}$, n = 1 $\ldots$  N}
    Obtain word vector ($\vec{wv_i}$), for each word $w_i$\;
    Calculate idf values, $idf(w_{i}),\,i = 1..|V|$ \tcc*{$|V|$ is vocabulary size}
    Cluster word vectors $\vec{wv}$ using GMM clustering into K clusters\;
    Obtain soft assignment $P(c_{k}|w_{i})$ for word $w_{i}$ and cluster $c_{k}$\; \tcc{Loop 5-10 can be pre-computed}
    
    \For{each word $w_i$ in vocabulary $V$}{
        \For{each cluster $c_k$}{
             $\vec{wcv_{ik}}$ $=$ $\vec{wv_i}$ $\times$ $P(c_{k}|w_{i})$\;
        }
        $\vec{wtv_{i}}$ $=$ $idf(w_i)$ $\times$ $\bigoplus_{k=1}^K$ $\vec{wcv_{ik}}$ \; \tcc{$\bigoplus$ is concatenation}
    }
    \For{ $n\in (1..N)$}{
        Initialize document vector $\vec{dv_{D_n}}$ = $\vec{0}$\;
        \For{word $w_i$ in $D_n$}{
            $\vec{dv_{D_n}}$ += $\vec{wtv_{i}}$\;
        }
        $\vec{SCDV_{D_n}}$ = make-sparse($\vec{dv_{D_n}}$)\;
        \tcc{as mentioned in sec \ref{SCDV}}
    }

\caption{Sparse Composite Document Vector}
\label{algo:SCDV}
\end{algorithm}

\subsection{Document Topic-vector Formation}
For each word $w_i$, we create $K$ different word-cluster vectors of d dimensions 
($\vec{wcv_{ik}}$) by weighing the word's embedding with its probability distribution in the  k$^{th}$ cluster, 
$P(c_{k}|w_{i})$. We then concatenate all K word-cluster vectors 
($\vec{wcv_{ik}}$) into a K$\times$d dimensional embedding and weigh it with inverse document frequency of 
$w_i$ to form a word-topics vector ($\vec{wtv_i}$). Finally, for all 
words appearing in document $D_n$, we sum their word-topic vectors 
$\vec{wtv_i}$ to obtain the document vector $\vec{dv_{D_n}}$.

\begin{equation*}
\vec{wcv_{ik}} = \vec{wv_i} \times P(c_{k}|w_{i})
\end{equation*}
\begin{equation*}
\vec{wtv_{i}} = idf(w_i) \times \bigoplus_{k=1}^K \vec{wcv_{ik}} 
\end{equation*}
where, $\bigoplus$ is concatenation

\subsection{Sparse Document Vectors}
\begin{figure}[h!]
\centering
\includegraphics[scale=0.35]{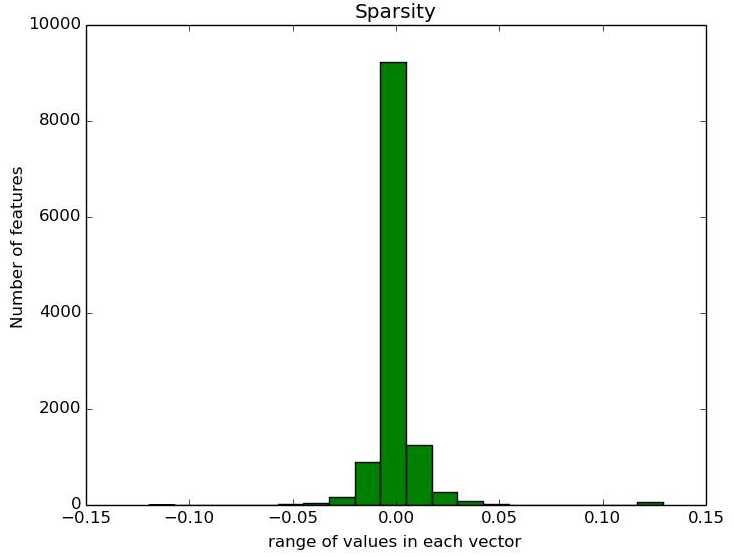}
\caption{ Distribution of attribute feature vector values.}
\label{figure:sparse}
\end{figure}

After normalizing the vector, we observed that most values in $\vec{dv_{D_n}}$ are very close to zero. Figure \ref{figure:sparse} verifies this observation. We utilize this fact to make the document vector 
$\vec{dv_{D_n}}$ sparse by zeroing attribute values whose absolute value 
is close to a threshold (specified as a parameter), which results in the Sparse Composite Document Vector $\vec{SCDV_{D_n}}$.

\begin{figure}[h!]
\centering
\includegraphics[scale=0.17]{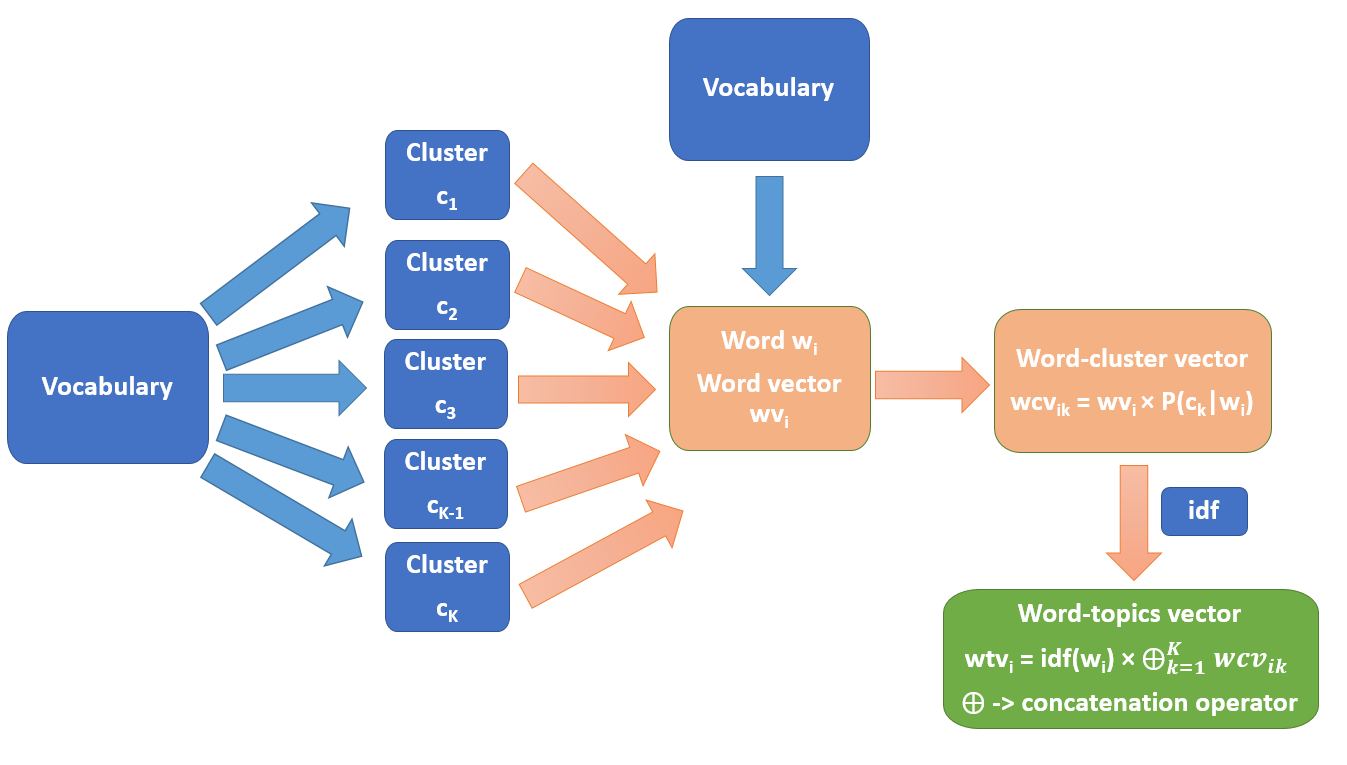}
\caption{Word-topics vector formation.}
\label{figure:wtv}
\end{figure}

\begin{figure}[h!]
\centering
\includegraphics[scale=0.17]{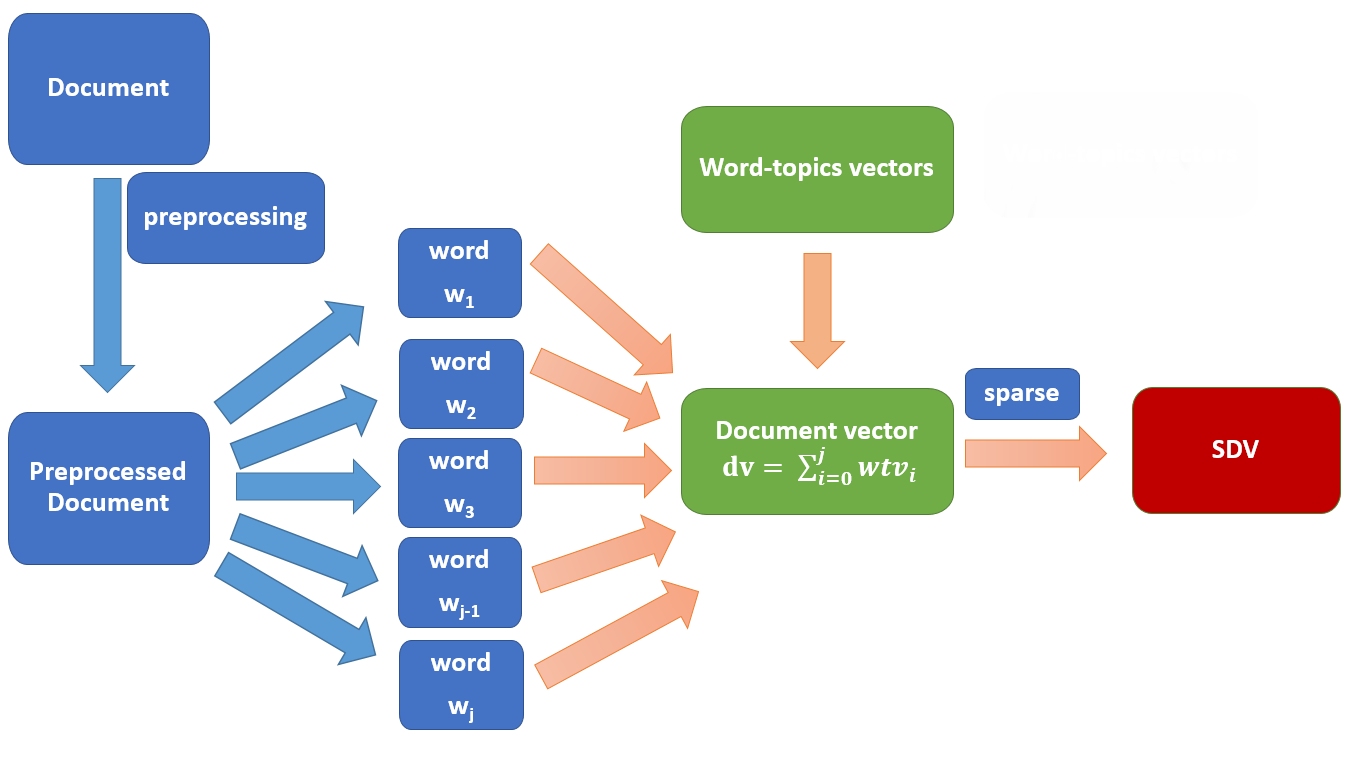}
\caption{Sparse Composite Document Vector formation.}
\label{figure:SCDV}
\end{figure}

In particular, let $p$ be percentage sparsity threshold parameter, $a_i$ 
the value of the $i^{th}$ attribute of the non-Sparse Composite Document Vector and $n$ represent the $n^{th}$ document in the training set:

\[
    a_i = 
\begin{cases}
    a_i & \text{if } {|a_i|}\geq {\frac{p}{100}}*t\\        0         & \text{otherwise}
\end{cases}
\]

\begin{equation*}
    t = \frac{|a_{min}| + |a_{max}|}{2}\\
\end{equation*}
\begin{equation*}
a_{min} = avg_{n}(min_{i}(a_i))
\end{equation*}
\begin{equation*}
a_{max} = avg_{n}(max_{i}(a_i))
\end{equation*}

Flowcharts depicting the formation of word-topics vector and Sparse Composite Document Vectors are shown in figure \ref{figure:wtv} and figure \ref{figure:SCDV} respectively. Algorithm \ref{algo:SCDV} describes SCDV in detail.

\section{Experiments}

We perform multiple experiments to show the effectiveness of SCDV representations for multi-class and multi-label text classification. For all experiments and baselines, we use Intel(R) Xeon(R) CPU E5-2670 v2 $@$ 2.50GHz, 40 working cores, 128GB RAM machine with Linux Ubuntu 14.4. However, we utilize multiple cores only during Word2Vec training and when we run the one-vs-rest classifier for Reuters.

\subsection{Baselines}

We consider the following baselines: Bag-of-Words (BoW) model  \cite{harris54}, Bag of  Word Vector (BoWV) \cite{vivek} model, paragraph vector models \cite{le2014distributed}, Topical word  embeddings (TWE-1)  \cite{AAAI159314}, Neural Tensor Skip-Gram Model (NTSG-1 to NTSG-3)  \cite{liu2015learning}, tf-idf weighted average word-vector model \cite{pranjal2015weighted} and weighted Bog of Concepts (weight-BoC)  \cite{boc}, where we build topic-document vectors by counting the member words in each topic.

We use the best parameter settings as reported in all our baselines to generate their results. We use $200$ dimensions for tf-idf weighted word-vector model, $400$ for paragraph vector model, $80$ topics and $400$ dimensional vectors for TWE, NTSG, LTSG and $60$ topics and $200$ dimensional word vectors for BOWV. We also compare our results with reported results of other topic modeling based document embedding methods like  $WTM$ \cite{wtm}, $w2v-LDA$ \cite{wtvlda}, $LDA$ \cite{liu2014topic}, $TV+MeanWV$ \cite{tvMeanWV}, $LTSG$ \cite{ltsg}, $Gaussian-LDA$ \cite{gaussianlda}, $Topic2Vec$  \cite{topic2vec},  \cite{lda2vec} and $MvTM$ \cite{mvtm}. Implementation of SCDV and related experiments is available here \footnote{https://github.com/dheeraj7596/SCDV}.

\subsection{Text Classification}
We run multi-class experiments on 20NewsGroup dataset \footnote{http://qwone.com/$\sim$jason/20Newsgroups/} and multi-label classification experiments on Reuters-21578 dataset \footnote{www.daviddlewis.com/resources/testcollections/reuters21578/}. We use the script\footnote{ https://gist.github.com/herrfz/7967781} for preprocessing the Reuters-21578 dataset. We use LinearSVM for multi-class classification and Logistic regression with OneVsRest setting for multi-label classification in baselines and SCDV.

For SCDV, we set the dimension of word-embeddings to $200$, sparsity threshold parameter to $4\%$ and the number of mixture components in GMM to $60$. All mixture components share the same spherical co-variance matrix. We learn word vector embedding using Skip-Gram with Negative Sampling (SGNS) of 10 and minimum word frequency as 20. We use 5-fold cross-validation on F1 score to tune parameter C of SVM.

\subsubsection{Multi-class classification}
We evaluate classifier performance using standard metrics like accuracy, macro-averaging precision, recall and F-measure. Table \ref{table:1} shows a comparison with the current state-of-art (NTSG) document representations on the 20Newsgroup dataset. We observe that SCDV outperforms all other current models by fair margins. We also present the class-wise precision and recall for 20Newsgroup on an almost balanced dataset with SVM over Bag of Words model and the SCDV embeddings in Table \ref{table:classwise} and observe that SCDV improves consistently over all classes.
\begin{table}[h!]
\caption{Performance on multi-class classification (Values in red show best performance, the SCDV algorithm of this paper)}
\vspace{0.2cm}
\begin{center}
\begin{tabular}{ |c|c|c|c|c| } 
 \hline
 {\bf Model} & {\bf Acc} & {\bf Prec} & {\bf Rec} & {\bf F-mes} \\
 \hline
  SCDV & \bf {\color{red}84.6} & \bf {\color{red}84.6} &\bf {\color{red}84.5}  &\bf {\color{red}84.6} \\
 NTSG-1 & 82.6 & 82.5 & 81.9 & 81.2\\
 NTSG-2 & 82.5 & 83.7 & 82.8 & 82.4\\
 BoWV & 81.6 & 81.1 & 81.1 & 80.9\\
 NTSG-3 & 81.9 & 83.0 & 81.7 & 81.1\\
 LTSG & 82.8 & 82.4 & 81.8 & 81.8\\
 WTM & 80.9 & 80.3 & 80.3 & 80.0\\
 w2v-LDA & 77.7 & 77.4 & 77.2 & 76.9\\
 TV+MeanWV & 72.2 & 71.8 & 71.5 & 71.6\\
MvTM & 72.2 & 71.8 & 71.5 & 71.6\\
 TWE-1 & 81.5 & 81.2 & 80.6 & 80.6\\
 lda2Vec & 81.3 & 81.4 & 80.4 & 80.5\\
 lda & 72.2 & 70.8 & 70.7 & 70.0\\
 weight-AvgVec &  81.9 & 81.7 & 81.9 & 81.7\\
 BoW & 79.7 & 79.5 & 79.0 & 79.0\\
 weight-BOC & 71.8 & 71.3 & 71.8 & 71.4\\
 PV-DBoW & 75.4 & 74.9 & 74.3 & 74.3 \\
 PV-DM & 72.4 & 72.1 & 71.5 & 71.5 \\ 
 \hline
\end{tabular}
\end{center}
\label{table:1}
\end{table}

\begin{table}[h!]
\caption{Class-level results on the balanced 20newsgroup dataset.}
\vspace{0.2cm}
\begin{center}
\begin{tabular}{ |c|c|c|c|c| }
\hline
\multicolumn{1}{|c|}{} & \multicolumn{2}{c|}{BoW} & \multicolumn{2}{c|}{SCDV} \\
 \hline
 {\bf Class Name} & {\bf Pre.} & {\bf Rec.} & {\bf Pre.} & {\bf Rec.} \\
 \hline
alt.atheism & 67.8 & 72.1 & 80.2 & 79.5 \\
comp.graphics & 67.1 & 73.5 & 75.3 & 77.4 \\
comp.os.ms-windows.misc & 77.1 & 66.5 & 78.6 & 77.2 \\
comp.sys.ibm.pc.hardware & 62.8 & 72.4 & 75.6 & 73.5 \\
comp.sys.mac.hardware & 77.4 & 78.2 & 83.4 & 85.5 \\
comp.windows.x & 83.2 & 73.2 & 87.6 & 78.6 \\
misc.forsale & 81.3 & 88.2 & 81.4 & 85.9 \\
rec.autos & 80.7 & 82.8 & 91.2 & 90.6 \\
rec.motorcycles & 92.3 & 87.9 & 95.4 & 95.7 \\
rec.sport.baseball & 89.8 & 89.2 & 93.2 & 94.7 \\
rec.sport.hockey & 93.3 & 93.7 & 96.3 & 99.2 \\
sci.crypt & 92.2 & 86.1 & 92.5 & 94.7 \\
sci.electronics & 70.9 & 73.3 & 74.6 & 74.9 \\
sci.med & 79.3 & 81.3 & 91.3 & 88.4 \\
sci.space & 90.2 & 88.3 & 88.5 & 93.8 \\
soc.religion.christian & 77.3 & 87.9 & 83.3 & 92.3 \\
talk.politics.guns & 71.7 & 85.7 & 72.7 & 90.6 \\
talk.politics.mideast & 91.7 & 76.9 & 96.2 & 95.4 \\
talk.politics.misc & 71.7 & 56.5 & 80.9 & 59.7 \\
talk.religion.misc & 63.2 & 55.4 & 73.5 & 57.2 \\
 \hline
\end{tabular}
\end{center}
\label{table:classwise}
\end{table}

\subsubsection{Multi-label classification}
We evaluate multi-label classification performance using Precision@K, nDCG@k \cite{bhatia2015sparse}, Coverage error, Label ranking average precision score (LRAPS)\footnote{Section $3.3.3.2$ of \\ $scikit-learn.org/stable/modules/model\_evaluation.html$} and F1-score. All measures are extensively used for the multi-label classification task. However, F1-score is an appropriate metric for multi-label classification as it considers label biases when train-test splits are random. Table \ref{table:4}  show evaluation results for multi-label text classification on the Reuters-21578 dataset.

\begin{table*}
\begin{center}
\caption{Performance on various metrics for multi-label classification for Reuters(Values in red show best performance, the SCDV algorithm of this paper)}
\vspace{0.3cm}
\label{table:4}
\begin{tabular}{ |c|c|c|c|c|c|c| } 
 \hline
{\bf Model} & \multicolumn{1}{|p{1.5cm}|}{\centering {\bf Prec@1} \\ {\bf nDCG@1}} & \multicolumn{1}{|p{1.1cm}|}{\centering {\bf Prec} \\ {\bf @5}} & \multicolumn{1}{|p{1.1cm}|}{\centering {\bf nDCG} \\ {\bf @5}} & \multicolumn{1}{|p{1.3cm}|}{\centering {\bf Coverage} \\ {\bf Error}} & {\bf LRAPS} & \multicolumn{1}{|p{1.5cm}|}{\centering {\bf F1-Score}}\\
 \hline
SCDV & \bf {\color{red}94.20} & \bf {\color{red}36.98} & \bf {\color{red}49.55} & \bf {\color{red}6.48} & \bf {\color{red}93.30}  & \bf {\color{red}81.75} \\
BoWV & 92.90 & 36.14 & 48.55 & 8.16 & 91.46 & 79.16 \\
TWE-1  & 90.91 & 35.49 & 47.54 & 8.16 & 91.46 & 79.16\\
PV-DM  & 87.54 & 33.24 & 44.21 &  13.15 & 86.21 & 70.24\\ 
PV-DBoW  & 88.78 &  34.51& 46.42 & 11.28 & 87.43 & 73.68\\
AvgVec  & 89.09 & 34.73 & 46.48 & 9.67 & 87.28 & 71.91\\
tfidf AvgVec & 89.33 & 35.04 & 46.83 & 9.42 & 87.90 & 71.97\\
\hline
\end{tabular}
\end{center}
\end{table*}

\subsubsection{Effect of Hyper-Parameters}
\begin{figure*}[h!]
  \includegraphics[scale=0.23]{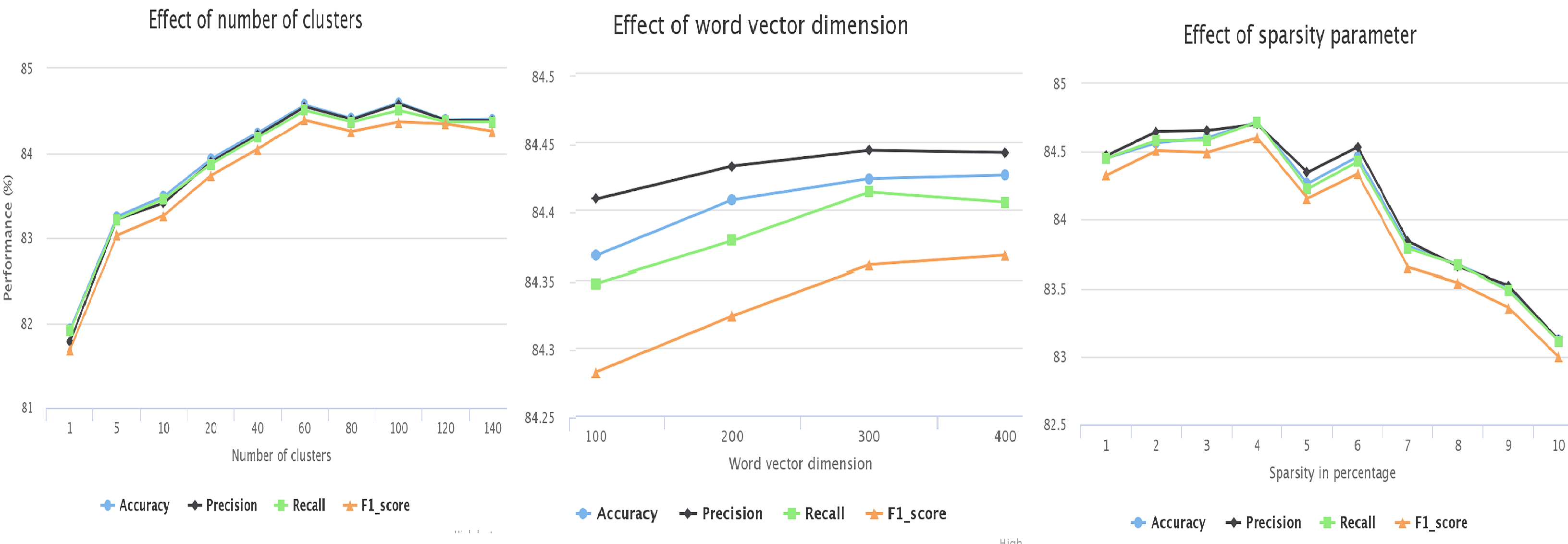}
  \caption{Effect of varying number of clusters (left), varying word vector dimension (center) and varying sparsity parameter (right) on performance for 20NewsGroup with SCDV}
  \label{figure:var_parameter}
\end{figure*}

SCDV has three parameters: the number of clusters, word vector dimension and sparsity threshold parameter. We vary one parameter by keeping the other two constant. Performance on varying all three parameters in shown in Figure \ref{figure:var_parameter}. We observe that performance improves as we increase the number of clusters and saturates at 60. The performance improves until a word vector dimension of 300 after which it saturates. Similarly, we observe that the performance improves as we increase $p$ till 4 after which it declines. At 4\% thresholding, we reduce the storage space by 80\% compared to the dense vectors.

\begin{figure*}[h!]
  \includegraphics[scale=0.16]{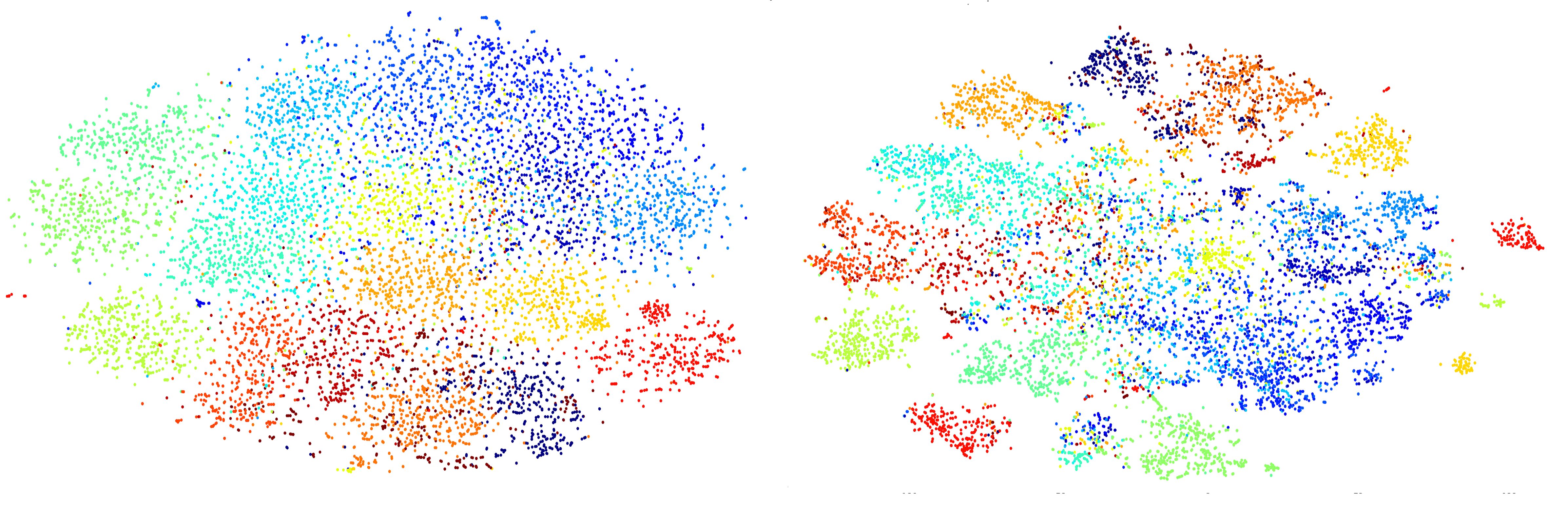}
  \caption{Visualization of paragraph vectors(left) and SCDV(right) using t-SNE}
  \label{figure:tsne}
\end{figure*}

\subsection{Topic Coherence}
We evaluate the topics generated by GMM clustering on 20NewsGroup for quantitative and qualitative analysis. Instead of using perplexity  \cite{chang2009reading}, which doesn't correlate with semantic coherence and human judgment of individual topics, we used the popular topic coherence  \cite{mimno2011optimizing}, \cite{arora2013practical}, \cite{liu2014topic} measure. A higher topic coherence score indicates a more coherent topic.

We used Bayes rule to compute the $P(w_{k}|c_{i})$ for a given topic $c_{i}$ and given word $w_{j}$ and compute the score of the top 10 words for each topic.
\begin{equation*}
    P(w_{k}|c_{i}) = \frac{P(c_{i}|w_{k}) P(w_{k})}{P(c_{i})}
\end{equation*}

where,
\begin{equation*}
    {P(c_{i})} = \sum_{i=1}^{K}{P(c_{i}|w_{k}) P(w_{k})}
\end{equation*}

\begin{equation*}
    P(w_{k}) = \frac{\#(w_{k})}{\sum_{i=1}^{V} \#(w_{i})}
\end{equation*}

Here, $\#(w_{k})$ denotes the number of times word $w_{k}$ appears in the corpus and V represents vocabulary size.

We calculated the topic coherence score for all topics for $SCDV$, $LDA$ and $LTSG$ \cite{ltsg}. Averaging the score of all 80 topics, GMM clustering
scores -85.23 compared to -108.72 of LDA and -92.23 of LTSG. Thus, SCDV creates more coherent topics than both LDA and LTSG.


\begin{table*}[t]
\small
\begin{center}
\caption{Top words of some topics from GMM and LDA on 20NewsGroup for K = 80. Higher score represent better coherent topics.}
\label{table:coherence}
\vspace{0.3cm}
\begin{tabular}{ |ccc|ccc|ccc|}
 \hline
 \multicolumn{3}{|c|}{Topic Image} & \multicolumn{3}{c|}{Topic Health} & \multicolumn{3}{c|}{Topic Mail} \\ \hline
 {\bf GMM } & {\bf LTSG} & {\bf LDA} & {\bf GMM } & {\bf LTSG} & {\bf LDA} & {\bf GMM } & {\bf LTSG} & {\bf LDA} \\ \hline
  file & image & image & heath & stimulation & doctor & ftp & anonymous & list\\
  bit & jpeg & file & study & diseases & disease & mail & faq & mail\\
  image & gif & color & medical & disease & coupons & internet & send & information\\
  files & format & gif & drug & toxin & treatment & phone & ftp & internet\\
  color & file & jpeg & test & toxic & pain & email & mailing & send\\
  format & files  & file & drugs & newsletter & medical & send & server& posting\\
  images & convert & format & studies & staff & day & opinions & mail & email\\
  jpeg & color & bit & disease & volume & microorganism & fax & alt & group\\
  gif & formats & images & education & heaths & medicine & address & archive & news\\
  program & images & quality & age & aids & body & box & email & anonymous\\
  \hline
  -67.16 & -75.66 & -88.79 & -66.91 & -96.98 & -100.39 & -77.47 & -78.23 & -95.47\\
  \hline
\end{tabular}
\end{center}
\end{table*}

Table \ref{table:coherence} shows top 10 words of 3 topics from $GMM$ clustering, $LDA$ model and $LTSG$ model on 20NewsGroup and $SCDV$ shows higher topic coherence. Words are ranked based on their probability distribution in each topic. Our results also support the qualitative results of  \cite{vec2topic} paper, where k-means was used over word vectors find topics.

\subsection{Context-Sensitive Learning}
In order to demonstrate the effects of soft clustering (GMM) during SCDV formation, we select some words (w$_{j}$) with multiple senses from 20Newsgroup and their soft cluster assignments to find the dominant clusters. We also select top scoring words (w$_{k}$) from each cluster (c$_{i}$) to represent the meaning of that cluster.
Table \ref{table:7} shows polysemic words and their dominant clusters with assignment probabilities. This indicates that using soft clustering to learn word vectors helps combine multiple senses into a single embedding vector.

\begin{table}[h]
\small
\begin{center}
\caption{Words with multiple senses assigned to multiple clusters with significant probabilities}
\label{table:7}
\begin{tabular}{|c|c|c|} 
 \hline
 {\bf Word } & {\bf Cluster Words} & {\bf P$(c_{i}|w_{j})$} \\
  \hline
  subject:1 & physics, chemistry, math, science & 0.27\\
  subject:2 & mail, letter, email, gmail & 0.72\\
   \hline
  interest:1 & information, enthusiasm, question & 0.65\\
  interest:2 & bank, market, finance,  investment & 0.32\\
  \hline
  break:1 & vacation, holiday, trip,  spring  & 0.52\\
  break:2 & encryption, cipher, security, privacy & 0.22\\
  break:3 & if, elseif, endif, loop, continue & 0.23\\
  \hline
  unit:1 & calculation, distance, mass, length & 0.25\\
  unit:2 & electronics, KWH, digital, signal & 0.69\\
  \hline
\end{tabular}
\end{center}
\end{table}

\subsection{Information Retrieval}

\citep{croft} used \cite{mikolov2013linguistic}'s paragraph vectors to enhance the basic language model based retrieval model. The language model(LM) probabilities are estimated from the corpus and smoothed using a Dirichlet prior \cite{zhailm}. In \citep{croft}, this language model is then interpolated with the paragraph vector (PV) language model as follows.
\begin{equation*}
    P(w|d) = (1-\lambda) P_{LM}(w|d) + \lambda P_{PV}(w|d)\\
\end{equation*}
where,
\begin{equation*}
	P_{PV}(w|d) = \frac{exp(\vec{w}.\vec{d})}{\sum_{i=1}^{V} exp(\vec{w_i}.\vec{d})} \\
\end{equation*}
and the score for document d and query string Q is given by
\begin{equation*}
	score(q,d) = \sum_{w \in Q}^{} P(w)P(w|d) \\
\end{equation*}
where $P(w)$ is obtained from the unigram query model and $score(q,d)$ is used to rank documents.
\cite{croft} do not directly make use of paragraph vectors for the retrieval task, but improve the document language model. To directly make use of paragraph vectors and make computations more tractable, we directly interpolate the language model query-document score $score(q,d)$ with the similarity score between the normalized query and document vectors to generate $score_{PV}(q,d)$, which is then used to rank documents.
\begin{equation*}
	score_{PV}(q,d) = (1-\lambda)score(q,d) +  \lambda \vec{q}.\vec{d}
\end{equation*}
Directly evaluating the document similarity score with the query paragraph vector rather than collecting similarity scores for individual words in the query helps avoid confusion amongst distinct query topics and makes the interpolation operation faster. In Table \ref{table:map}, we report Mean Average Precision(MAP) values for four datasets, Associated Press 88-89 (topics 51-200), Wall Street Journal (topics 51-200), San Jose Mercury (topics 51-150) and Disks 4 \& 5 (topics 301-450) in the TREC collection. We learn $\lambda$ on a held out set of topics. We observe consistent improvement in MAP for all datasets. We marginally improve the MAP reported by \cite{croft} on the Robust04 task. In addition, we also report the improvements in MAP score when Model based relevance feedback \cite{zhaimb} is applied over the initially retrieved results from both models. Again, we notice a consistent improvement in MAP. 

\begin{table*}[h!]
\caption{ Mean average precision (MAP) for IR on four IR datasets}
\label{table:map}
\begin{center}
\begin{tabular}{ |c|c|c|c|c| } 
\hline
{\bf Dataset} & {\bf LM} & {\bf LM+SCDV} & {\bf MB} & { \bf MB + SCDV }\\
\hline
AP & 0.2742 & {\color{red}0.2856} & 0.3283 & {\color{red}0.3395}\\ 
SJM & 0.2052 & {\color{red}0.2105} & 0.2341 & {\color{red}0.2409}\\
WSJ & 0.2618 &  {\color{red}0.2705} & 0.3027 & {\color{red}0.3126}\\
Robust04 & 0.2516 & {\color{red}0.2684} & 0.2819 & {\color{red}0.2933}\\
\hline
\end{tabular}
\end{center}
\end{table*}

\begin{table}[h!]
\caption{ Time Comparison (20NewsGroup) (Values in red show least time, the SCDV algorithm of this paper) }
\begin{center}
\begin{tabular}{ |c|c|c|c| } 
 \hline
 {\bf Time (sec)} & {\bf BoWV} & {\bf TWE-1} & {\bf SCDV}\\
 \hline
 DocVec Formation & 1250 & 700 &\bf {\color{red}160}\\ 
 Total Training & 1320 & 740  &\bf {\color{red}200}\\
 Total Prediction & 780 & 120  &\bf {\color{red}25}\\
 \hline
\end{tabular}
\end{center}
\label{table:2}
\end{table}

\section{Analysis and Discussion}
SCDV overcomes several challenges encountered while training document vectors, which we had mentioned above.

%

\begin{enumerate}
\item Clustering word-embeddings to discover topics improves performance of classification as Figure \ref{figure:var_parameter} (left) indicates, while also generating coherent clusters of words (Table \ref{table:coherence}). Figure \ref{figure:tsne} shows that clustering gives more discriminative representations of documents than paragraph vectors do since it uses K $\times$ d dimensions while paragraph vectors embed documents and words in the same space. This enables SCDV to represent complex documents. Fuzzy clustering allows words to belong to multiple topics, thereby recognizing polysemic words, as Table \ref{table:7} indicates. Thus it mimics the word-context interaction in NTSG and LTSG.
\item Semantically different words are assigned to different topics. Moreover, a single document can contain words from multiple different topics. Instead of a weighted averaging of word embeddings to form document vectors, as most of the previous work does, concatenating word embeddings for each topic (cluster) avoids merging of semantically different topics. 
\item It is well-known that in higher dimensions, structural regularizers such as sparsity help overcome the curse of dimensionality  \cite{structuredsparcity}.Figure \ref{figure:sparse} demonstrates this, since majority of the features are close to zero. Sparsity also enables linear SVM to scale to large dimensions. On 20NewsGroups, BoWV model takes up 1.1 GB while SCDV takes up only 236MB($~80\%$ decrease). Since GMM assigns a non-zero probability to every topic in the word embedding, noise can accumulate when document vectors are created and tip the scales in favor of an unrelated topic. Sparsity helps to reduce this by zeroing out very small values of probability.
\item SCDV uses Gaussian Mixture Model (GMM) while $TWE$, $NTSG$ and $LTSG$ use LDA for finding semantic topics respectively. GMM time complexity is  $\mathcal{O}(VNT^{2})$ while that of LDA is $\mathcal{O}(V^{2}NT)$. Here, V = Vocabulary size, N = number of documents and T = number of topics. Since number of topics T $<$ vocabulary size V, GMM is faster. Empirically, compared to $TWE$, $SCDV$ reduces document vector formation, training and prediction time significantly. Table \ref{table:2} shows training and prediction times for BoWV, SCDV and TWE models.
\end{enumerate}

\section {Conclusion}
In this paper, we propose a document feature formation technique for topic-based document representation. SCDV outperforms state-of-the-art models in multi-class and multi-label classification tasks. SCDV introduces sparsity in document vectors to handle high dimensionality. Table \ref{table:2} indicates that SCDV shows considerable improvements in feature formation, training and prediction times for the 20NewsGroups dataset. We show that fuzzy GMM clustering on word-vectors lead to more coherent topic than LDA and can also be used to detect Polysemic words. SCDV embeddings also provide a robust estimation of the query and document language models, thus improving the MAP of language model based retrieval systems. In conclusion, SCDV is simple, efficient and creates a  more
accurate semantic representation of documents.

\section*{Acknowledgments}
The authors wants to thank Nagarajan Natarajan (Post-Doc, Microsoft Research, India),  Praneeth Netrapalli (Researcher, Microsoft Research, India), Raghavendra Udupa (Researcher, Microsoft Research, India), Prateek Jain (Researcher, Microsoft Research, India) for encouraging and valuable feedback .


\bibliography{eacl2017}
\bibliographystyle{eacl2017}


\end{document}